\pdfoutput=1

\documentclass[11pt]{article}

\usepackage[preprint]{acl}

\usepackage{times}
\usepackage{latexsym}

\usepackage[T1]{fontenc}

\usepackage[utf8]{inputenc}

\usepackage{microtype}

\usepackage{inconsolata}

\usepackage{graphicx}

\usepackage{hyperref}       
\usepackage{url}            
\usepackage{booktabs}       
\usepackage{amsfonts}       
\usepackage{nicefrac}       
\usepackage{microtype}      
\usepackage{xcolor}         
\usepackage{float}
\usepackage{wrapfig}  
\usepackage{tikz}  
\usepackage{parskip}

\usepackage{algorithm,algpseudocode}
\usepackage{algorithmicx}

\usepackage{amsmath}     
\usepackage{amssymb}       
\usepackage{amsfonts}     
\usepackage{appendix}
\usepackage{algorithm} 
\usepackage{algpseudocode} 

\usepackage{caption}

\newcommand{\ourmethod}{\textit{LoRASC}}

\usepackage[capitalize]{cleveref}
\crefname{figure}{Fig.}{Figs.}
\crefname{section}{Sec.}{Secs.}
\Crefname{section}{Section}{Sections}
\Crefname{table}{Table}{Tables}
\crefname{table}{Tab.}{Tabs.}

\title{Expressive and Generalizable Low-rank Adaptation for Large Models \\ via Slow Cascaded Learning}

\author{Siwei Li\textsuperscript{1}\thanks{Equal contribution. This work was done during Siwei Li's internship at Microsoft Research Asia.}, Yifan Yang\textsuperscript{2}\footnotemark[1]\thanks{Corresponding author.}, Yifei Shen\textsuperscript{2}, Fangyun Wei\textsuperscript{2}, Zongqing Lu\textsuperscript{1}, Lili Qiu\textsuperscript{2}, Yuqing Yang\textsuperscript{2} \\
  \textsuperscript{1}Tsinghua University, \textsuperscript{2}Microsoft Research Asia \\
  \texttt{siweili505@outlook.com, yifanyang@microsoft.com}}

\begin{document}
\maketitle

\begin{abstract}
Efficient fine-tuning plays a fundamental role in modern large models, with low-rank adaptation emerging as a particularly promising approach. However, the existing variants of LoRA are hampered by limited expressiveness, a tendency to overfit, and sensitivity to hyperparameter settings. This paper presents LoRA Slow Cascade Learning (\ourmethod), an innovative technique designed to enhance LoRA's expressiveness and generalization capabilities while preserving its training efficiency. Our approach augments expressiveness through a cascaded learning strategy that enables a mixture-of-low-rank adaptation, thereby increasing the model's ability to capture complex patterns. Additionally, we introduce a slow-fast update mechanism and cascading noisy tuning to bolster generalization. The extensive experiments on various language and vision datasets, as well as robustness benchmarks, demonstrate that the proposed method not only significantly outperforms existing baselines, but also mitigates overfitting, enhances model stability, and improves OOD robustness. Code will be release in \url{https://github.com/microsoft/LoRASC} very soon.



\end{abstract}
\section{Introduction}
Foundation models, which are large-scale models pre-trained on extensive datasets and subsequently adapted for specific downstream tasks, have become integral to contemporary machine learning frameworks. Fine-tuning these models is essential, yet full parameter fine-tuning often encounters significant memory and computational bottlenecks. As a result, Parameter-Efficient Fine-Tuning (PEFT) techniques, which aim to minimize the number of trainable parameters to reduce training costs and improve training stability, have gained increasing prominence. Among these techniques, Low-Rank Adaptation (LoRA)~\cite{hu2021lora} stands out due to its efficiency in reducing training costs through low-rank approximation for full-parameter updates. However, despite LoRA's advantages, its limitations in terms of expressiveness and generalization have been noted. Some studies suggest that the inherent low-rankness of LoRA might restrict its expressiveness~\cite{xia2024chain,meng2024periodiclora,relora,huang2024mixture}, with a preference for overparameterization, while others indicate a tendency for LoRA to overfit or exhibit overconfidence~\cite{lin2024lora,wang2023lora}.

In this work, we investigate the potential of cascading learning to augment the expressiveness of LoRA. Our approach involves initializing a new LoRA module at the start of each epoch and integrating this module into the backbone network after the epoch concludes. By employing a mixture-of-low-rank adaptation, we effectively increase the model's rank, while maintaining low training costs, as each cascading step consumes no more parameters and memory than a single LoRA model. Moreover, this method does not add any inference overhead by remerging each LoRA module into the backbone network.

To improve LoRA's generalization capabilities, we draw inspiration from optimization techniques. We repurpose certain strategies from optimizers for LoRA, motivated by the observation that initializing a new LoRA module for each epoch can represent a descent direction for the dataset. In optimization theory, flat minimizers are preferred, as they are associated with better generalization \cite{hochreiter1997flat,keskar2016large}. Inspired by the fact that the moving average mechanism guides models towards flat minimizers \cite{SWA}, we maintain both fast-updating and its moving average version, the slow-updating LoRA experts. The fast-updating expert is reinitialized regularly to learn from the data over a set number of steps, while the slow-updating expert undergoes updates via a proportional exponential moving average after the fast-updating cycle completes. Additionally, mirroring techniques in deep learning optimizers where noise proportional to the gradient scale is used to find flat minima \cite{xie2020diffusion}, we introduce noise at the beginning of each epoch, with the scale tied to the norm of LoRA's weights. 

To verify the effectiveness of the proposed method, we conduct extensive experiments on both language and vision tasks. For language tasks, we utilized the Llama2 model on $12$ datasets (e.g., SuperGLUE, SQuAD, DROP, GSM8K, and InstructEval), Alpaca among other instruct following benchmarks to demonstrate the effectiveness of our design. We can directly apply our approach to LoRA, LoRA+~\cite{hayou2024lora+}, Dora~\cite{liu2024dora}, and other members of the LoRA family, significantly improving their performance in large model transfer learning. For vision tasks, we also validated our approach on the CLIP pre-trained Vit-bigG model with the ImageNet dataset, showing a significant performance improvement relative to LoRA on domain adaptation datasets such as Image-R and Image-C. The proposed method consistently outperforms the baselines by a large margin.

\section{Related Work}
\label{related}

\subsection{Low-Rank Adaptation Finetuning}
Low-Rank Adaptation(LoRA)~\cite{hu2021lora} is a parameter-efficient fine-tuning method designed to adapt large models to new tasks, demonstrating superior performance. LoRA+~\cite{hayou2024lora+} improves performance and fine-tuning speed by setting different learning rates for the LoRA adapter matrices A and B with a carefully chosen ratio, maintaining the same computational cost as LoRA. Dora~\cite{liu2024dora} decomposes the pre-trained weight into two components, magnitude and direction, for fine-tuning, specifically employing LoRA for directional updates to efficiently minimize the number of trainable parameters. Our work introduces a robust cascading learning schedule for various LoRA variants, proving through extensive experiments that it can enhance the training performance of LoRA, LoRA+, and Dora without additional training costs.

\subsection{Combination of LoRA}
LoRAhub~\cite{huang2023lorahub} presents a simple framework designed for the purposeful assembly of LoRA modules trained on diverse tasks, aiming to achieve adaptable performance on unseen tasks. MOLE~\cite{huang2024mixture} treats each layer of trained LoRAs as a distinct expert and implements hierarchical weight control by integrating a learnable gating function within each layer. LoRAFlow~\cite{wang2024loraflow} utilizes dynamic weights to adjust the impact of different LoRAs. These methods are not in conflict with $\ourmethod$, as they focus on learning the combination of LoRA experts across different domains, while our method aims to learn more generalizable experts within a single domain using slow cascade learning.

ReLoRA~\cite{relora} enhances LoRA’s fitting ability by continuously merging online LoRA into the main network and restarting optimizer parameters during training. It also proposes a jagged cosine scheduler to implement a learning rate resume strategy at each step. COLA~\cite{xia2024chain} explores a similar approach but in a simpler manner, merely restarting optimizer parameters when initializing new LoRAs without adjusting the learning rate schedule. Our work employs a simpler cascading learning strategy where each expert learns independently for each epoch, without additional design for learning schedules or optimizer parameters. Additionally, we incorporate noise tuning and slow-fast update strategy, ensuring robustness in each expert merged into the pre-trained model. Our method can be applied to various LoRA variants, demonstrating effectiveness across multiple tasks in both language and image domains.

\section{Methods}

\subsection{LoRA}
Low-Rank Adaptation (LoRA) is a parameter-efficient fine-tuning method designed to adapt large pre-trained models to specific tasks with significantly fewer trainable parameters. Instead of updating all parameters of the model, LoRA inserts low-rank matrices into each layer of the pre-trained model, which are then fine-tuned. This reduces the computational burden and the risk of overfitting.

Given a pre-trained weight matrix $W_0 \in \mathbb{R}^{d \times k}$ in a neural network, LoRA approximates the update $\Delta W$ using two low-rank matrices $A \in \mathbb{R}^{d \times r}$ and $B \in \mathbb{R}^{r \times k}$, where $r \ll \min(d, k)$. The update is defined as:
\begin{equation}
    \Delta W = B A
\end{equation}
During fine-tuning, instead of updating $W$, we update $A$ and $B$, which results in:
\begin{equation}
    W = W_0 + \Delta W = W_0 + B A
\end{equation}
This low-rank adaptation significantly reduces the number of trainable parameters from $d \times k$ to $r \times (d + k)$.

\begin{figure*}[ht]
    \centering
    \includegraphics[width=\textwidth]{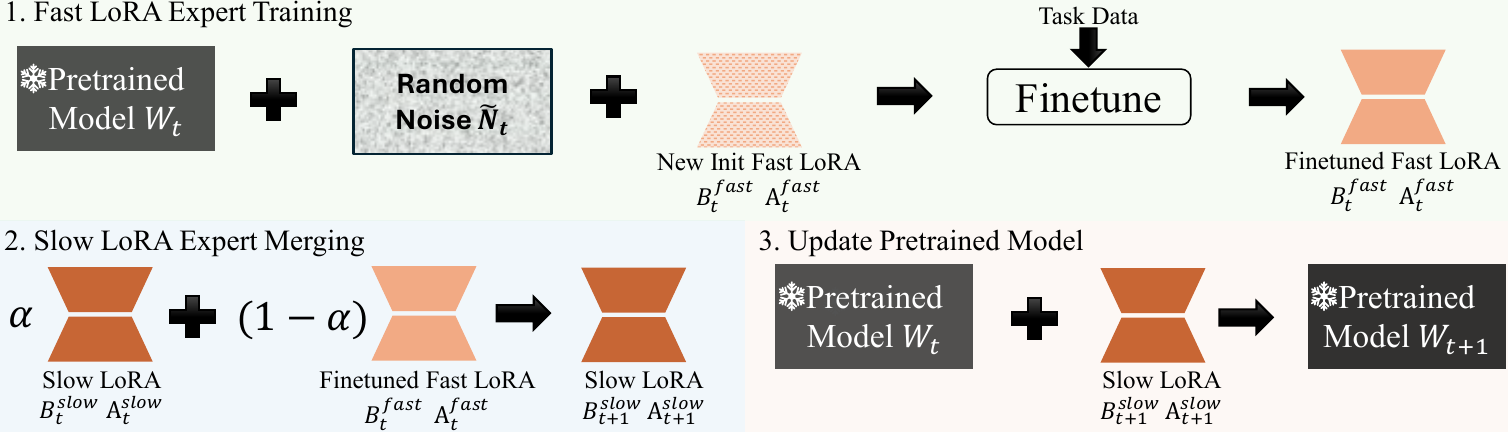}
    \caption{Iterative pipeline of $\ourmethod$. Here, \(t\) represents the iteration step, and \(BA\) denotes the low-rank learnable vectors in LoRA. The backbone network \(W\) always has its gradients turned off, and \(\alpha\) is the hyperparameter controlling the pace of the slow-fast update. Our method follows three stages: 1. Fast LoRA expert training, where noise is added to the backbone network, followed by training the fast LoRA on the task data. 2. Slow LoRA expert merging, where a portion of the learned fast LoRA is weighted and merged into the slow LoRA. 3. Update the pretrained model, merging the updated slow LoRA into the backbone network, and prepare for the next iteration.}
    \label{fig:architecture}
\end{figure*}

\subsection{\ourmethod{}}
\subsubsection{Cascading LoRA Learning}
\label{clora}
Due to the reparameterization nature of low-rank adaptation (LoRA) fine-tuning, employing multiple LoRA experts incurs the same inference cost as using a single LoRA expert. This characteristic makes LoRA particularly suitable for integration with cascading learning to enhance performance in transfer learning tasks. As analyzed in ReLoRA~\cite{relora}, reinitializing new LoRA modules during the learning schedule can progressively increase the model's rank, thereby improving its fitting ability.

In $\ourmethod$, we default to learning one LoRA expert per epoch. After training one LoRA expert, it is merged into the main network, and the next expert learns based on the optimized residuals. The optimization schedule for each single LoRA expert is a compressed version of the original full-training schedule: for instance, if a model was originally trained for $N$ epochs, each expert in $\ourmethod$ completes training in 1 epoch with fixed starting and ending learning rates with the same but compressed scheduler. This makes $\ourmethod$ easy to apply to any large model transfer learning scenario using LoRA, without requiring changes to hyperparameters. The only necessary adjustment is an increase in the learning rate. Since the number of training steps is compressed, each step must be larger to cover the same distance. Additionally, Li et al. \cite{li2019towards} found that higher learning rates can lead to stronger generalization ability, which might also explain the improved out-of-domain performance of our method.

Mathematically, the cascading LoRA learning can be described as follows:

1. For each epoch $t$, train a new LoRA expert $(A_t, B_t)$ to minimize the residual error, where $\mathcal{L}$ is the fine-tuning loss function:
\begin{equation}
(A_t, B_t) = \arg \min_{A_t, B_t} \mathcal{L} \left( W_{t-1} + B_t A_t \right),
\end{equation}

2. Merge the trained LoRA expert into the main network:
\begin{equation}
    W_{t} = W_{t-1} + B_t A_t
\end{equation}

By iteratively merging each new LoRA expert into the main network, loRA cascading progressively enhances the model's capacity to fit the data without increasing the inference cost.

\subsubsection{LoRA Slow-Fast Update}
To enhance the generalization of large model transfer learning, we aim to avoid local optima at each step of cascading. Even with low-rank adaptation, this issue persists due to the imbalance between model parameters and training data. Inspired by SWA~\cite{SWA}, which averages model parameters over several epochs to find a more generalized solution, we employ a sliding average method to ensure the stability and robustness of each LoRA merged into the main network.

Specifically, during training, we maintain two LoRA experts at each cascading step $t$ as shown in ~\Cref{fig:architecture}: a slow-updating LoRA $(A^{\text{slow}}_t, B^{\text{slow}}_t)$ and a fast-updating LoRA $(A^{\text{fast}}_t, B^{\text{fast}}_t)$. At step 0, both slow and fast LoRA share the same initialization. During each cascading iteration, fast LoRA undergoes fine-tuning, and after completion, it is averaged with slow LoRA. The slow LoRA is then merged into the pre-trained model, while the fast-updating LoRA is reinitialized for the next iteration. We control the retention proportion of the slow expert with a hyperparameter $\alpha$.

The update rules are given by:

\begin{equation}
    A^{\text{slow}}_{t+1} = \alpha A^{\text{slow}}_t + (1 - \alpha) A^{\text{fast}}_t
\end{equation}
\begin{equation}
    B^{\text{slow}}_{t+1} = \alpha B^{\text{slow}}_t + (1 - \alpha) B^{\text{fast}}_t
\end{equation}

By employing this slow-fast update strategy, $\ourmethod$ ensures that each merged LoRA expert contributes to a more generalized solution, enhancing the overall stability and performance of the model in transfer learning scenarios.




\subsubsection{Cascading Noisy Tuning}
To further enhance generalization, we introduce random noise to the pre-trained model before each new LoRA fine-tuning step. Unlike NoisyTune~\cite{wu2022noisytune}, which adds uniform noise to different parameter matrices according to their standard deviations only once at the beginning of fine-tuning, we apply noise before training each new expert. This approach helps the model escape local optima at every slow LoRA step, thereby reducing the risk of overfitting.

\begin{algorithm}[h]
\caption{Pseudo Code for \ourmethod}
\label{alg:ourmethod}
\begin{algorithmic}[1]
\Require Pre-trained model weights $W_0$, number of epochs $T$, loss function $\mathcal{L}$, slow update parameter $\alpha$, noise parameter $\lambda$
\State Initialize $W \gets W_0$
\State Initialize $A^{\text{slow}}$, $B^{\text{slow}}$ \Comment{Initialize slow LoRA matrices}
\State Initialize $A^{\text{fast}} \gets A^{\text{slow}}$, $B^{\text{fast}} \gets B^{\text{slow}}$ \Comment{Fast LoRA matrices initialized from slow ones}
\For{epoch $t = 1$ to $T$}
    \If{$t > 1$}
        \State Reinitialize $A^{\text{fast}}$, $B^{\text{fast}}$ \Comment{Reinitialize fast LoRA matrices for subsequent epochs}
    \EndIf
    \State $\widetilde{W} \gets W + U\left(-\frac{\lambda}{2}, \frac{\lambda}{2}\right) \cdot \text{std}(B^{\text{slow}} A^{\text{slow}})$
    \State optimizer $\gets$ \text{InitializeOptimizer}($A^{\text{fast}}, B^{\text{fast}}$)
    \State lr\_scheduler $\gets$ \text{InitializeLRScheduler}(optimizer)
    \For{batch in training data}
        \State Forward pass: $L \gets \mathcal{L}(\widetilde{W} + B^{\text{fast}} A^{\text{fast}})$
        \State Backward pass: \text{Compute gradients}
        \State optimizer.step()
        \State lr\_scheduler.step()
    \EndFor
    \State \text{Update slow LoRA}: 
    \State $A^{\text{slow}} \gets \alpha A^{\text{slow}} + (1 - \alpha) A^{\text{fast}}$
    \State $B^{\text{slow}} \gets \alpha B^{\text{slow}} + (1 - \alpha) B^{\text{fast}}$
    \State \text{Merge slow LoRA into main network}: $W \gets \widetilde{W} + B^{\text{slow}} A^{\text{slow}}$
\EndFor
\State \Return $W$
\end{algorithmic}
\end{algorithm}

Additionally, the presence of the slow-updating LoRA module indicates the direction of parameter changes under the new task. Therefore, we use the standard deviation of the slow LoRA weights to determine the noise scale rather than the pre-trained model's weights. Incorporating this noise before every expert ensures that the model continuously explores robust and flatten parameter spaces, thus improving generalization and reducing the tendency to overfit.

The perturbation is defined as:
\begin{equation}
    \widetilde{{N}}_t = U\left(-\frac{\lambda}{2}, \frac{\lambda}{2}\right) \cdot \text{std}(B^{\text{slow}}_{t} A^{\text{slow}}_{t})
\end{equation}

where std stands for standard deviation. The function $U(a, b)$ represents uniform distribution noise
ranged from $a$ to $b$, and $\lambda$ is a hyperparameter that controls the relative noise intensity.

\subsection{Overview}
With LoRA cascading learning, slow-fast updates and noisy tuning, the pipeline of our $\ourmethod$ is as follows:
\begin{equation}
    \widetilde{W}_{t-1} = W_{t-1} + \widetilde{{N}}_t 
\end{equation}

\begin{equation}
    (A^{\text{fast}}_t, B^{\text{fast}}_t) = \arg \min_{A^{\text{fast}}_t,B^{\text{fast}}_t } \mathcal{L} \left( \widetilde{W}_{t-1} + B^{\text{fast}}_t A^{\text{fast}}_t \right)
\end{equation}

\begin{equation}
    A^{\text{slow}}_{t} = \alpha A^{\text{slow}}_{t-1} + (1 - \alpha) A^{\text{fast}}_t
\end{equation}
\begin{equation}
    B^{\text{slow}}_{t} = \alpha B^{\text{slow}}_{t-1} + (1 - \alpha) B^{\text{fast}}_t
\end{equation}
\begin{equation}
    W_{t} = \widetilde{W}_{t-1} + B^{\text{slow}}_{t} A^{\text{slow}}_{t}
\end{equation}

$\ourmethod$ pipeline can be seen in~\Cref{fig:architecture}. Although we use vanilla LoRA to show slow casdade learning, $\ourmethod$ should be able to boost the performance of any LoRA variants, such as DoRA~\cite{liu2024dora}, LoRA+~\cite{hayou2024lora+}, LoRA-FA~\cite{lorafa}, etc. Moreover, $\ourmethod$ is easy to implement, and we provide pseudocode with more detailed explanations in ~\Cref{alg:ourmethod}.

\section{Experiments}
\label{headings}

We conducted extensive experiments to demonstrate the effectiveness and robustness of $\ourmethod$ across both NLP and CV domains. 

For language tasks, we conducted our language experiments using the popular open-source large language model, Llama2\footnote{\url{https://huggingface.co/meta-llama/Llama-2-7b-hf}}. We evaluated our approach on several NLU and GLU tasks, selecting both SuperGLUE~\cite{wang2019superglue} tasks (including classification and multiple-choice ) and generation tasks. We also tested the model's performance in mathematical reasoning using the GSM8K dataset~\cite{cobbe2021training}. Additionally, we performed instruction tuning experiments to verify the transfer learning capability of our method, achieving significant improvements on key metrics such as MMLU~\cite{hendrycks2020measuring}, DROP~\cite{dua-etal-2019-drop}, BBH~\cite{srivastava2022beyond} and HumanEval~\cite{chen2021evaluating}. 

For visual tasks, we chose the CLIP ViT-bigG/14\footnote{\url{https://huggingface.co/laion/CLIP-ViT-bigG-14-laion2B-39B-b160k}} as our pretrained model, fine-tuning it on the ImageNet-1K~\cite{deng2009imagenet} training set and testing it on the validation set. Subsequently, we evaluated the trained model on perturbed datasets such as ImageNet-A~\cite{hendrycks2021natural}, ImageNet-C~\cite{hendrycks2019benchmarking}, ImageNet-R~\cite{hendrycks2021many}, ImageNet-V2~\cite{recht2019imagenet}, ImageNet-Sketch~\cite{wang2019learning} and Stylized-ImageNet ~\cite{geirhos2018imagenet} demonstrating our method's robustness and generalization capabilities.

\subsection{Implementation Details}

For all experiments, we exclusively fine-tuned $q$ and $v$ in attention layers as delineated by ~\citet{malladi2023fine} and ~\citet{ren2024mini}. The fine-tuning process utilized single NVIDIA H100 GPU. For all tasks, we explored several learning rates and reported the optimal performance. For the hyper-parameters of $\ourmethod$, we explored the factor $\alpha$ of Slow-Fast Update in \{0.5, 0.6, 0.8\} to control the updating ratio. Additionally, we selected the noise intensity from \{0.1, 1, 10\}, which is a significantly smaller set compared to the default 7 in NoistTune~\cite{wu2022noisytune}. All the results were averaged across 3 distinct random seeds, and we report the optimal performance.

\begin{table*}[t]
\centering
\setlength{\tabcolsep}{1pt}
\small
\resizebox{\textwidth}{!}{
    \begin{tabular}{lcccccccccccc}
    \toprule
    Task & \textbf{SST-2} & \textbf{RTE} & \textbf{CB} & \textbf{BoolQ} & \textbf{WSC} & \textbf{WIC} & \textbf{MultiRC} & \textbf{COPA} & \textbf{ReCoRD} & \textbf{SQuAD} & \textbf{DROP} & \textbf{GSM8K} \\
    Task type & \multicolumn{7}{c}{---------------- classification ----------------} & \multicolumn{2}{c}{-- multiple choice --} & \multicolumn{2}{c}{-- generation --} & \multicolumn{1}{c}{- math -} \\
    \midrule
    \textbf{LoRA} & 95.5 & 87.4 & 91.1 & 85.7 & 70.2 & 72.4 & 85.3 & 85.0 & 81.2 & 90.4 & 51.6 & 19.5 \\
     w/ \textbf{COLA} & 95.9 & 87.7 & 91.1 & 85.7 & 66.4 & 72.6 & 85.3 & 82.0 & 81.4 & 90.6 & 51.6 & 21.0 \\
    \textit{w/ \textbf{\ourmethod}} & & & & & & & & & & & & \\
    \hspace{3mm}+ Cascade & 95.8 & 87.7 & 92.9 & 86.1 & 71.1 & 72.3 & 86.3 & \textbf{88.0} & 81.6 & 91.8 & 52.5 & 21.5 \\
    \hspace{3mm}++ Slow LoRA & 96.0 & 88.0 & 96.4 & 86.8 & 74.0 & 72.1 & 86.3 & \textbf{88.0} & 82.1 & 92.7 & 55.3 & \textbf{27.5} \\
    \hspace{3mm}+++ Noise Tuning & \textbf{96.1} & \textbf{88.1} & \textbf{96.5} & \textbf{87.4} & \textbf{75.0} & \textbf{72.7} & \textbf{86.6} & \textbf{88.0} & \textbf{82.2} & \textbf{92.9} & \textbf{56.7} & \textbf{27.5} \\
   
    \midrule
    \textbf{LoRA+} & 95.7 & 87.0 & 91.4 & 85.9 & 69.2 & 72.1 & 85.7 & 87.0 & 81.3 & 90.5 & 55.8 & 22.0 \\
    \textit{w/ \textbf{\ourmethod}} & & & & & & & & & & & & \\
    \hspace{3mm}+ Cascade & 95.7 & 87.0 & \textbf{92.9} & 86.2 & 71.2 & 72.8 & 85.3 & \textbf{88.0} & \textbf{81.9} & 91.2 & 55.8 & 19.5 \\
    \hspace{3mm}++ Slow LoRA & 95.7 & \textbf{88.1} & \textbf{92.9} & 85.9 & 67.3 & 73.5 & 85.7 & \textbf{88.0} & \textbf{81.9} & \textbf{92.0} & 56.3 & 23.0 \\
    \hspace{3mm}+++ Noise Tuning & \textbf{95.8} & \textbf{88.1} & \textbf{92.9} & \textbf{86.3} & \textbf{71.4} & \textbf{74.1} & \textbf{86.1} & \textbf{88.0} & \textbf{81.9} & \textbf{92.0} & \textbf{56.4} & \textbf{24.0} \\
    \midrule
    \textbf{DoRA} & 95.4 & 87.4 & 96.4 & 85.7 & 72.1 & 71.5 & 84.7 & 88.0 & 81.1 & 91.1 & 54.8 & 21.0 \\
    \textit{\textbf{w/ \ourmethod}} & & & & & & & & & & & & \\
    \hspace{3mm}+ Cascade & 95.8 & 87.4 & 96.4 & 85.8 & 65.4 & \textbf{72.8} & 84.1 & 88.0 & 81.6 & 91.7 & 52.6 & 22.5 \\
    \hspace{3mm}++ Slow LoRA & 95.8 & 88.1 & 96.4 & 85.8 & 65.4 & \textbf{72.8} & 86.1 & 88.0 & 81.9 & 92.8 & 54.8 & 25.0 \\
    \hspace{3mm}+++ Noise Tuning & \textbf{96.0} & \textbf{88.5} & \textbf{96.5}& \textbf{87.6} & \textbf{75.6} & \textbf{72.8} & \textbf{86.8} & \textbf{89.0} & \textbf{82.2} & \textbf{93.3} & \textbf{56.5} & \textbf{25.5} \\
    \bottomrule
    \end{tabular}
}
\caption{Comparative Performance of LoRA, LoRA+, and DoRA enhanced with $\ourmethod$ across multiple in-domain fine-tuning datasets. }
\label{tab:main}
\end{table*}

\begin{table*}[t]
\centering
\resizebox{0.6\textwidth}{!}{  
    \begin{tabular}{lccccc}
        \toprule
        Method & \textbf{MMLU} & \textbf{DROP } & \textbf{HEval } & \textbf{BBH }  &\textbf{GSM8K}\\
        \midrule
        \textbf{LoRA} & 45.83& 32.76& 31.26& 13.41
 &11.5\\
        \textit{w/ \textbf{\ourmethod}} & & & &  &\\
        \hspace{3mm}+ Cascade & 45.53& 32.71& \textbf{31.61}& 14.02
 &11.5\\
        \hspace{3mm}++ Slow LoRA & 45.68& \textbf{33.74}& 31.38& \textbf{17.07}
 &12.5\\
        \hspace{3mm}+++ Noise & \textbf{45.98}& 33.02& \textbf{31.61}& 15.24
 &\textbf{16.5}\\
        \bottomrule
    \end{tabular}
}
\caption{Results on instruction-following tasks. The model was trained on Alpaca and evaluated on InstructEval metrics and GSM8K. \ourmethod{} consistently achieves the best performance compare to vanilla LoRA.}
\label{tab:sft}
\end{table*}

\begin{figure*}[ht]
    \centering
    \includegraphics[width=\textwidth]{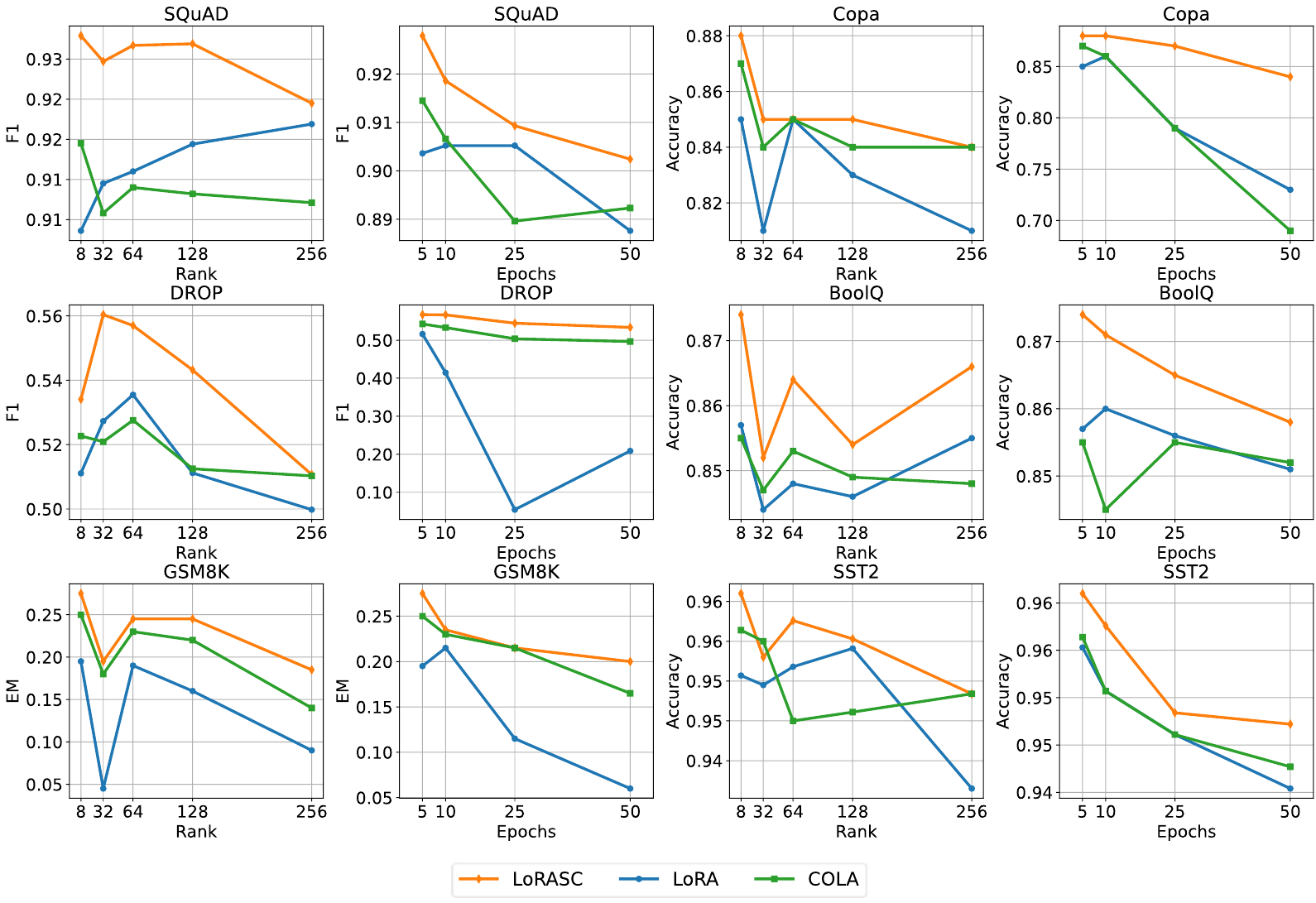}
    \caption{Performance of $\ourmethod$ compared to LoRA and COLA  across various ranks and learning schedules in a subset of text transfer learning tasks. It can be observed that $\ourmethod$ consistently achieves stable performance improvements across all ranks and learning schedules, particularly at higher ranks and longer epochs, where $\ourmethod$ can mitigate performance degradation caused by overfitting.}
    \label{fig:ep_rank}
\end{figure*}

\subsection{Main Results}
\subsubsection{\ourmethod{} for Large Language Model}

\paragraph{Experiment setting. }
For in-domain language transfer learning, we consider the SuperGLUE dataset collection ~\cite{wang2019superglue}, including: BoolQ~\cite{clark-etal-2019-boolq},
CB~\cite{de2019commitmentbank}, COPA~\cite{roemmele2011choice},  MultiRC~\cite{khashabi2018looking}, ReCoRD~\cite{zhang2018record}, RTE~\cite{socher2013recursive}, WiC~\cite{pilehvar-camacho-collados-2019-wic}, and WSC~\cite{levesque2012winograd}.
We also include SST-2~\cite{dagan2005pascal} , GSM8K~\cite{cobbe2021training}  and two question answering(QA) datasets, SQuAD~\cite{rajpurkar-etal-2016-squad} and DROP~\cite{dua-etal-2019-drop}. And we directly used 8-shot direct prompting or GSM8K evaluation\footnote{\url{https://github.com/allenai/open-instruct}}. We adhered to the experimental configuration described by \citet{malladi2023fine}, randomly selecting 1000 examples for training, 
500 for validation, and 1000 for testing across each dataset. The AdamW optimizer was employed, 
with training spanning 5 epochs, consistent with the baseline settings. A linear learning rate 
schedule was implemented, with the initial learning rate selected from \{1×10\textsuperscript{-5}, 5×10\textsuperscript{-5}, 1×10\textsuperscript{-4}, 5×10\textsuperscript{-4}, 1×10\textsuperscript{-3}\}. By default the batch size was set to 4 and the LoRA rank
was set to 8. For LoRA+, we adhered
to its setup by fixing the learning rate of $B$ matrices to be 16 times that of $A$ matrices. DoRA decomposes the pre-trained weight into magnitude and direction components, with LoRA efficiently updating the direction component. This means that each LoRA expert represents DoRA's direction component. When applying $\ourmethod$ to DoRA, we maintain continuous training of the magnitude while applying our technique to the direction component. We follow the standard procedure of merging and reinitializing LoRA and align it with the slow-fast update and noisy tuning. 

For instruction tuning, we use the Alpaca\footnote{\url{https://github.com/tatsu-lab/stanford_alpaca/}}~\cite{alpaca} dataset for training. The batch size was set to 128. We follow the training scripts of~\citet{ren2024mini} in our experiment. We finetune our model for 3 epochs. A linear learning rate schedule was applied, with the initial learning rate selected from \{1×10\textsuperscript{-4}, 3×10\textsuperscript{-4}, 5×10\textsuperscript{-4}, 1×10\textsuperscript{-3}\}. For evaluation we use InstructEval\footnote{\url{https://github.com/declare-lab/instruct-eval}}~\cite{chia2023instructeval}, 5-shot direct prompting for MMLU , 3-shot direct prompting for BBH and DROP, 0-shot direct prompting for HEval.

\paragraph{$\ourmethod$ exhibits excellent adaptability to LoRA variants.}

In the experiments shown in~\Cref{tab:main}, $\ourmethod$ outperforms the COLA across various tasks, demonstrating the effectiveness of our LoRA cascading technique. Moreover, $\ourmethod$ effectively boosted the performance of LoRA, LoRA+, and DoRA across 12 in-domain training datasets encompassing four major tasks: classification, multiple choice, generation, and mathematics. $\ourmethod$ achieved significant improvements across all these tasks, demonstrating its ability to enhance the learning capabilities and in-domain generalization of the LoRA family of models. Moreover, the progressive addition of cascading learning, slow-fast updates, and noisy tuning further improved performance, validating the design of our approach. The robust slow cascading strategy not only enhanced overall performance but also provided strong generalization capabilities.

\paragraph{$\ourmethod$ on Instruction-Following tasks.}

\Cref{tab:sft} presents the performance of our proposed method, $\ourmethod$, applied to LoRA across several instruction-following tasks. These instruction-following tasks are particularly challenging due to the weak correlation between the training data and the benchmarks, making them entirely out-of-domain tests. Despite this difficulty, our method achieved notable improvements across various evaluation metrics used in InstructEval and GSM8K. Furthermore, the design of slow-fast updates and noisy tuning still steadily enhanced the performance of cascading learning, further validating the effectiveness of our approach and motivation.


\begin{table*}
    \centering
    \resizebox{0.7\textwidth}{!}{
    \begin{tabular}{lccccccc} 
         \toprule
         Method & \textbf{ImageNet } & \textbf{IN-V2} & \textbf{IN-C } & \textbf{IN-R } & \textbf{IN-A } & \textbf{IN-SK } & \textbf{IN-ST } \\ 
         \midrule
         \textbf{LoRA} & 87.1 & 77.7& 66.2& 87.1& 72.6& 64.9 & 24.1\\
         \textit{w/ \textbf{\ourmethod}} & & & & & & & \\
         \hspace{3mm}+ Cascade & 87.1& 77.5 & 66.7& 88.5& \textbf{73.6}& 65.4& 24.3\\
         \hspace{3mm}++ Slow LoRA & 87.7& 78.3& \textbf{66.8}& 88.1& 73.4& 65.2& 24.1\\
         \hspace{3mm}+++ Noise Tuning & \textbf{87.8}& \textbf{78.4}& \textbf{66.8}& \textbf{88.7}& 73.4& \textbf{65.5}& \textbf{24.4} \\
         \bottomrule
    \end{tabular}
    }
    \caption{Top-1 accuracy of various methods on ImageNet-1K and 6 robustness benchmarks. The table compares the baseline LoRA with our three proposed techniques. Our approach demonstrates improved robustness on the ViT-bigG model across all the evaluated benchmarks.}
    \label{tab:vit}
\end{table*}

\begin{table*}
    \centering
    \setlength{\tabcolsep}{1pt}
    \resizebox{\columnwidth}{!}{
    \begin{tabular}{cccccccc} 
         \toprule
         Experts & \textbf{RTE} & \textbf{DROP}   & \textbf{WIC} & \textbf{BoolQ} &\textbf{ReCoRD} &\textbf{SST-2} &\textbf{SQuAD} \\ 
         \midrule
         2& 87.0& 53.8& 72.4& 85.3&81.3& 95.5&92.0\\
         5&\textbf{88.1}& \textbf{56.7}& \textbf{72.6}& \textbf{87.4}&\textbf{82.2}& \textbf{96.1}&\textbf{92.9}\\
 25& 86.7& 51.2& 70.5& 83.5& 81.4& 95.5&92.2\\
         125& 83.8& 50.2& 70.5& 84.5&81.2& 95.1&90.7\\
         1250&83.8& 49.4& 69.4& 85.3&81.1& 92.9&88.1\\
         \bottomrule
    \end{tabular}
    }
    \caption{Evaluation with varing expert number of \ourmethod . The highest average performance for each task is highlighted
in bold.}
    \label{expnum}
\end{table*}

\subsubsection{$\ourmethod$ for CLIP ViT-bigG}
\paragraph{Experiment setting. }
For the ImageNet-1K visual classification task, to validate the transfer performance of our method on larger vision models, we selected CLIP ViT-bigG/14 as our pre-training backbone.We utilized the AdamW optimizer and a cosine scheduler, training for a total of 10 epochs on the ImageNet-1K training set. The batch size was fixed at 64, and the learning rate was chosen from \{1×10\textsuperscript{-4}, 5×10\textsuperscript{-4}, 1×10\textsuperscript{-3}\}. 
For evaluation, we first test our model on the ImageNet-1K validation set using top-1 accuracy. To demonstrate the improvement in our method's transferability and robustness, we conducted further tests on robustness benchmarks from ~\citet{mao2022easyrobust} for transfer learning tasks. 

\paragraph{Evaluation of $\ourmethod$ on ImageNet and Robustness Benchmarks. }

Table \ref{tab:vit} showcases the performance of our proposed method,  $\ourmethod$, applied to LoRA on ImageNet-1K and several robustness benchmarks, including IN-V2, IN-C, IN-R, IN-A, IN-SK, and IN-ST. These benchmarks test the model's robustness and generalization ability beyond the standard ImageNet dataset. Our method demonstrates consistent improvements in top-1 accuracy across all evaluated benchmarks.  $\ourmethod$ consistently enhances the robustness and generalization of the ViT-bigG model across these challenging benchmarks, validating the effectiveness of cascading learning, slow-fast updates, and noisy tuning in improving model performance in diverse and robust scenarios.


\subsection{Ablation Study and Analysis}

\paragraph{Larger Ranks and Longer Epochs.} As shown in ~\Cref{fig:ep_rank}, $\ourmethod$ consistently achieves more stable performance on datasets such as SQuAD, DROP, and GSM8K compared to both LoRA and COLA, which also employs a cascading strategy. This validates our motivation: $\ourmethod$ is a training strategy that retains LoRA’s beneficial properties while seamlessly enhancing its fitting ability and robust generalization.

\paragraph{Ablation for $\ourmethod$ Expert Cascade Frequency.} $\ourmethod$ defaults to updating once per epoch, as each expert completes training on the entire dataset within one epoch. In ~\Cref{expnum}, we experimented with different update frequencies. In this setting, we trained for a total of 5 epochs, with each epoch consisting of 250 iterations, resulting in a total training period of 1250 iterations. The table shows that having 5 experts, corresponding to one new expert per epoch, yields the optimal performance. Interestingly, we observe that even with 1250 experts, where a new expert is initialized every iteration, the model still achieves highly competitive performance. In this extreme case, following ~\Cref{alg:ourmethod}, the model cannot iterate the learning rate as each backpropagation step is immediately followed by the initialization of a new expert. We speculate that the strong generalization capability of slow cascading compensates for the weak fitting ability in this scenario. With 2 experts(one expert every 2.5 epochs), which aligns with COLA's default setting for this scenario, the performance is lower than $\ourmethod$'s default of one expert per epoch. This may be due to the model being more prone to local optima after 2.5 epochs, which negatively impacts the effectiveness of slow cascading.

\section{Conclusion}
In this paper, we address the limitations of fine-tuning large pre-trained models, particularly the issue of overfitting and the high computational costs associated with transferring these models to niche tasks. We introduce a novel technique, $\ourmethod$, which enhances the Low-Rank Adaptation (LoRA) approach by integrating cascading learning, slow-fast updates, and noisy tuning. Our method aims to improve the fitting capability and generalization of LoRA models without incurring additional computational costs.

We provide a detailed analysis of $\ourmethod$ and demonstrate its effectiveness through extensive experiments in both the natural language processing (NLP) and computer vision (CV) domains. Our method consistently outperforms baseline LoRA models and their variants (LoRA+, Dora) across multiple datasets and tasks, including SuperGLUE, SQuAD, DROP, GSM8K, and various instruction-following benchmarks. Additionally, our method enhances the robustness and transferability of vision models on ImageNet and several robustness benchmarks.

\section*{Limitations}
While \ourmethod{} attempts to find a better balance between model convergence and generalization, it does not fundamentally resolve the issue. Our proposed mechanisms of slow-fast updating and noisy tuning can enhance model generalization and prevent overfitting; however, if the magnitude of these adjustments is too large, it may still lead to difficulties in model convergence. Therefore, it is necessary to adjust the $\alpha$ parameter in the slow-fast merging process and $\lambda$ in the intensity of noise added to each expert according to the specific task. In our experiments, only a few candidate adjustments were needed to significantly outperform vanilla LoRA, yet this still incurs additional costs. Adaptive adjustment of these parameters according to the task is a direction for future work that we intend to explore. 

Additionally, while this study only explores LoRA cascading learning for single training tasks and finds it to effectively enhance model performance, in practice, we could combine LoRA experts from multiple domains, similar to the MoLE~\cite{huang2024mixture} approach, to further improve model capabilities. In such cases, how to better perform slow cascading would be an interesting issue to address.


\bibliography{egbib}




\end{document}